\documentclass[10pt,twocolumn,letterpaper]{article}
\usepackage{cvpr}
\usepackage{times}
\usepackage{epsfig}
\usepackage{graphicx}
\usepackage{amsmath}
\usepackage{amssymb}

\usepackage[utf8]{inputenc} 
\usepackage[T1]{fontenc}    
\usepackage{url}            
\usepackage{booktabs}       
\usepackage{amsfonts}       
\usepackage{nicefrac}       
\usepackage{microtype}      
\usepackage[numbers]{natbib}

\usepackage{hyperref}

\hypersetup{
  colorlinks,
  breaklinks=true,
  anchorcolor={blue!50!black},
  linkcolor={blue!50!black},
  citecolor={blue!50!black},
  urlcolor={blue}
}
\usepackage{xcolor}

\cvprfinalcopy 

\begin{document}

\title{Unsupervised Deep Metric Learning via Auxiliary Rotation Loss}

\author{
Xuefei Cao\\
Brown University\\
{\tt\small xuefei\_cao@brown.edu}
\and
Bor-Chun Chen\\
University of Maryland\\
{\tt\small sirius@umd.edu}
\and
Ser-Nam Lim\\
Facebook AI\\
{\tt\small sernam@gmail.com}
}

\maketitle

\begin{abstract}
Deep metric learning is an important area due to its applicability to many domains such as image retrieval and person re-identification. 
The main drawback of such models is the necessity for labeled data. 
In this work, we propose to generate pseudo-labels for deep metric learning directly from clustering assignment and we introduce unsupervised deep metric learning (UDML) regularized by a self-supervision (SS) task. In particular, we propose to regularize the training process by predicting image rotations.
Our method (UDML-SS) jointly learns discriminative embeddings, unsupervised clustering assignments of the embeddings, as well as a self-supervised pretext task. UDML-SS iteratively cluster embeddings using traditional clustering algorithm (e.g., k-means), and sampling training pairs based on the cluster assignment for metric learning, while optimizing self-supervised pretext task in a multi-task fashion. The role of self-supervision is to stabilize the training process and encourages the model to learn meaningful feature representations that are not distorted due to unreliable clustering assignments. The proposed method performs well on standard benchmarks for metric learning, where it outperforms current state-of-the-art approaches by a large margin and it also shows competitive performance with various metric learning loss functions.
\end{abstract}

\begin{figure*}[h!]
\begin{center}
\includegraphics[width=0.95\textwidth]{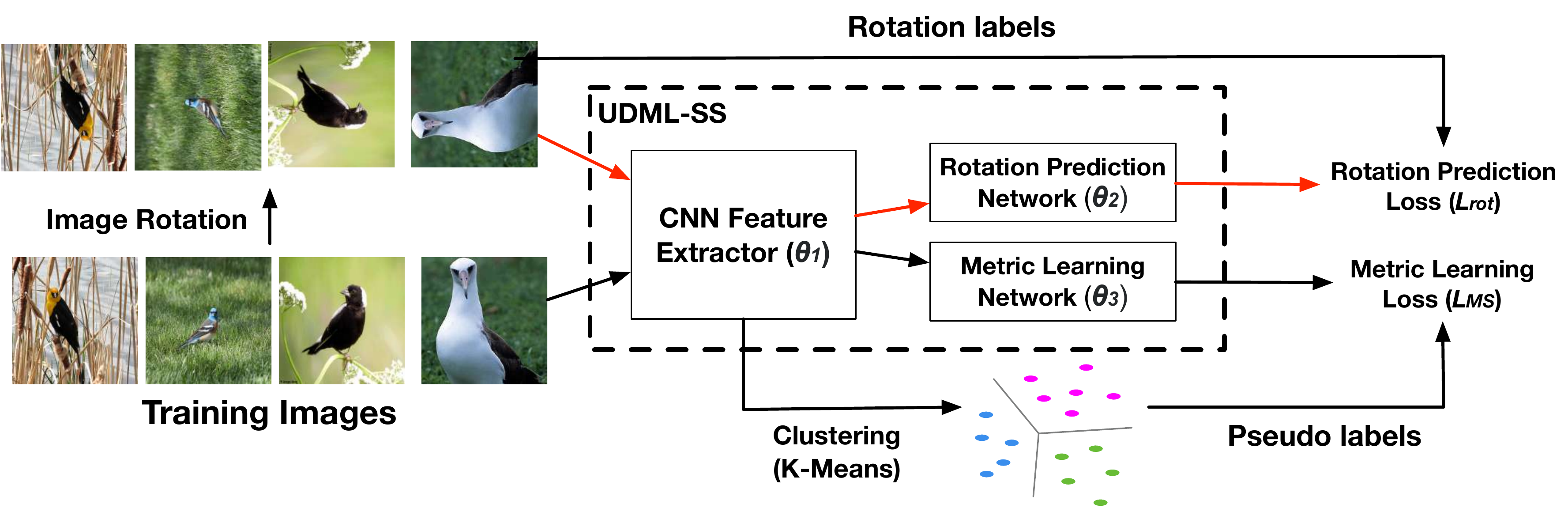}
\end{center}
  \caption{Unsupervised metric learning with rotation-based self-supervision. The red arrows indicate the network flow for the rotation prediction task.  For the rotation loss, all images are rotated by 0, 90, 180, and 270 degrees. and images are classified by the fully connected layer according to their rotation degree. 
  The black arrows indicate the network flow for the metric learning loss. 
  The pseudo-labels are initiated by the cluster assignment of features extracted from pre-trained convnet. After the first iteration, the features will be extracted directly from the fully connected layer of metric learning loss. We learn the parameters of neural networks, and the cluster assignments of the resulting embedding vectors iteratively.
  }
\label{fig:diagram}
\end{figure*}

\section{Introduction}

Metric learning methods aim to learn effective embedding space where similar instances are mapped to nearby points, while for samples coming from different classes, the embedding vectors are pushed apart. These methods explore different loss functions and mining methods to measure the similarities between data points accurately and robustly. Unlike traditional classification tasks which focus on category-specific concepts, metric learning aims to learn the general concept of distance metrics ~\cite{oh2016deep}. With the recent success of deep neural networks in computer vision, deep metric learning methods have shown impressive results. Deep metric learning methods have applications in different domains, such as person re-identification ~\cite{hermans2017defense, yu2018hard}, image retrieval ~\cite{wohlhart2015learning, he2018triplet, grabner20183d}, near-duplicate detection ~\cite{zheng2016improving}, zero-shot learning ~\cite{bucher2016improving, bucher2016hard,yelamarthi2018zero} and visual tracking ~\cite{hu2015deep,leal2016learning}.
However, to obtain better performance, the training process often requires large-scale labeled data. Most of the fine-grained datasets are especially expensive to annotate since annotators are required to be domain experts ~\cite{gebru2017fine}. Thus unsupervised deep metric learning is becoming of great interest to the vision community. 

A major goal of unsupervised representation learning is to learn similarities between images or weak category information without labeled instances ~\cite{ye2019unsupervised}. Recent work ~\cite{bautista2016cliquecnn,bautista2017deep,caron2018deep} treats the classification problem as a pretext task and explores the idea of updating the weights of models by predicting the cluster assignments. Iscen \etal ~\cite{iscen2018mining} introduce a fully unsupervised way to mine hard training samples. Ye \etal ~\cite{ye2019unsupervised} propose to sample positive pairs by using data augmentation and treat different instances as negative pairs.

The idea of using cluster assignment as pseudo-labels has been studied in deep learning domain ~\cite{bautista2016cliquecnn, bautista2017deep, caron2018deep, caron2019unsupervised} such as DeepCluster ~\cite{caron2018deep}. However, these studies mainly focus on classification tasks with pseudo-labels. 
In UDML-SS, we propose to use pseudo-labels directly to generate samples for metric learning loss. By sampling the positive and negative pairs based on the cluster assignments generated with k-means, we are able to update weights of the unsupervised metric learning model. The pseudo-labels are then re-computed given the new embedding vectors and this process iterates until the model converges.

A challenge with clustering is that it tends to contain many unreliable assignments, which causes instability during training and difficulty of converges. Motivated by the recent development of self-supervised learning, we mitigate this problem by adding an auxiliary, self-supervised loss to the metric learning loss. 
This leads to more stable and accurate training because the dependency of the learned representations on the quality of the
clustering assignment is reduced. In particular, we apply the state-of-the-art self-supervision method based on image rotation ~\cite{gidaris2018unsupervised}.

\textbf{Our contributions} \quad 
In this work, we present an unsupervised metric learning framework (UDML-SS). We propose a metric learning loss that is based on cluster assignments directly as well as combines the metric learning with self-supervised representation learning. 
It alternates between clustering the learned embedding vectors
and updating the weights of the convnet by minimizing a loss function, which is a combination of metric learning loss and a self-supervised loss. 
For simplicity, we focus our study on k-means for the former. For the latter, the image rotation prediction task ~\cite{gidaris2018unsupervised} is chosen as our self-supervision task. 
UDML-SS is conceptually simple and compatible with any metric learning loss functions, which we will show in Section \ref{exp}. Our method is evaluated extensively on several benchmarks for metric learning, where it outperforms current state-of-the-art unsupervised metric learning approaches by a large margin, e.g.,
improving ~\cite{ye2019unsupervised} by +8.5\% Recall@1 on
CUB200 ~\cite{wah2011caltech}, by +3.8\% Recall@1 on Cars196 ~\cite{krause20133d} and by +14.6\% Recall@1 on Stanford Online Product (Product) ~\cite{oh2016deep}. Figure \ref{fig:diagram} shows a conceptual pipeline of the proposed approach.



\section{Related Work}
\textbf{Metric Learning}\quad  With the
progress made in deep learning, many approaches have been proposed for supervised deep metric learning. A lot of research effort has been devoted to designing new loss functions.
Classical pair-based loss functions including contrastive loss ~\cite{hadsell2006dimensionality, hu2014discriminative} and triplet loss ~\cite{schroff2015facenet, cheng2016person} are widely used in most existing metric
learning methods. Contrastive loss encourages samples from a
positive pair to be closer, and maximizes the distance between a negative pair in the embedding space. Triplet loss defines
each triplet by choosing a positive sample and a
negative sample given the same anchor point. It
aims to learn an embedding where the similarity of the
negative sample plus a given margin is lower than that of the positive one to the anchor. Extended from triplet loss, quadruplets are also applied in recent work, such as histogram loss ~\cite{ustinova2016learning}. Other methods, such as lifted-structure ~\cite{oh2016deep}, n-pair
loss ~\cite{sohn2016improved}, angular loss ~\cite{wang2017deep}, adapted triplet loss ~\cite{yu2018correcting},  multi-similarity ~\cite{wang2019multi} focus on fully utilizing pairwise relations of all points in a batch. 
Hard sample mining has also been widely adopted to produce more robust models. 
Here, instead of sampling all negative instances for an anchor point, the most challenging negative instances are mined.
To this end, Schroff \etal ~\cite{schroff2015facenet} propose semi-hard mining. They sample a negative example within the batch, such that it is close to the anchor point but further away from positives. Wu \etal ~\cite{wu2017sampling} improve it by uniformly sampling negative
instances weighted by their distance. Ge \etal ~\cite{ge2018deep} introduce a new violate
margin, which is computed dynamically over the constructed hierarchical tree. 
Duan \etal ~\cite{duan2018deep} introduce a deep adversarial metric learning framework to generate synthetic hard negatives
from the observed negative samples.
To fully exploit information buried in all samples,  Zheng \etal ~\cite{zheng2019hardness} performs linear interpolation on embeddings to adaptively manipulate their hard levels so that the metric is always challenged with proper difficulty.  All these metric learning methods are supervised with class labels.

\textbf{Self-supervised Representation Learning} 
Self-supervised representation learning has been widely used in different domains ~\cite{doersch2015unsupervised, lee2017unsupervised,jang2018grasp2vec}.
Self-supervised representation learning utilizes only
unlabeled data to formulate a pretext learning task
for which a target objective can be acquired without supervision.
~\cite{doersch2015unsupervised, noroozi2016unsupervised} predict the relative position of image patches to learn semantically relevant content. 
Larsson \etal~\cite{larsson2016learning} use colorization as a proxy task. Giaris \etal~\cite{gidaris2018unsupervised} propose to rotate
the image and predict the rotation angle, which is a simple but yet effective method to achieve useful representations for downstream
image classification and segmentation tasks. Feng \etal~\cite{feng2019self}  introduce a split representation that contains both rotation related and unrelated part. ~\cite{oord2018representation, henaff2019data, bachman2019learning} propose to
train feature extractors by maximizing an estimate of the mutual information (MI)
between different views of the data. Although these methods show state of the art performance on the classification task, it is unclear whether MI maximization is a good objective for learning good representations in an unsupervised fashion ~\cite{tschannen2019mutual}.
 Recently, Hendrycks ~\etal~\cite{hendrycks2019using} show self-supervised representation learning can improve the robustness of the classification model to label corruption. 

\textbf{Deep Clustering} \quad Clustering is a popular unsupervised learning method. Caron \etal~\cite{caron2018deep} proposes a scalable clustering approach for the unsupervised representation learning of visual features. It iterates between clustering with k-means the features generated by the deep nets and using a discriminative loss to update the parameters by predicting the cluster assignments as pseudo-labels. In ~\cite{caron2018deep, caron2019unsupervised}, deep clustering idea is explored for general unsupervised feature learning, where the main goal is to pre-train model without labels.

\textbf{Unsupervised Metric learning} \quad Most of the metric learning methods are supervised with class labels. There have been relatively fewer efforts devoted to unsupervised metric learning. 
~\cite{bautista2016cliquecnn, bautista2017deep} 
split the training set into different groups based on complicated clustering scheme and utilize induced classification problem as a pretext task. Iscen \etal
~\cite{iscen2018mining} introduce an unsupervised framework
for hard training example mining which exploits the manifold distance to extract hard examples.
Ye \etal~\cite{ye2019unsupervised} instead aim at
learning data augmentation invariant features and explore the instance-wise supervision. This method is related to another unsupervised learning method ~\cite{wu2018unsupervised}.

\section{Proposed Method}

\subsection{Problem Formulation}
Let $\mathcal{X}$ denote the data space where we sample a set of
unlabeled data points $X = [x_1, x_2, ... , x_n]$. Let $f_{\theta_1} : \mathcal{X} \to \mathcal{W}$ be a mapping from the data
space to a feature space, where we have $w_i = f_{\theta_1}(x_i)$. $f$ is usually represented by a convolutional
neural network (CNN), e.g. the pre-trained Inception-V1 ~\cite{szegedy2015going} on
ImageNet. Mapping $f_{\theta_1}$ learns a non-linear transformation of the image into a deep feature space $\mathcal{W}$.
The objective of metric learning is to learn a metric
in the feature space so that it can measure the visual similarity correctly based on different datasets. 
To learn the mapping from feature space to the embedding space, another function $g_{\theta_2} : \mathcal{W} \to \mathcal{Z}$ is appended to project feature vectors to embedding vectors. 
The embedding vector $g_{\theta_2}(w_i)$ is usually normalized to have a unit length for training stability ~\cite{schroff2015facenet}. 
Finally, two mappings $f_{\theta_1}$ and $g_{\theta_2}$ are jointly learned (where the feature extraction backbone is usually fine-tuned) in such a way that $g_{\theta_2} \circ f_{\theta_1}$ maps images within same categories (positive pairs)
close to one another and images in different categories (negative pairs)
far apart in the embedding space. 
The similarity between two data points in the embedding space is thus defined as 
\begin{equation}
    S(x_i, x_j) = \langle g_{\theta_2}(f_{\theta_1}(x_i)), g_{\theta_2}(f_{\theta_1}(x_j)) \rangle.
\end{equation}
Supervised metric learning approaches would use labeled data points to construct a training set ($\mathcal{T}$) of
positive and negative pairs of items. And the network parameters are learned by minimizing a specific loss function:
\begin{equation}
    \theta_1, \theta_2 = \operatorname*{argmin}_{\theta_1, \theta_2} L(\mathcal{T}, \theta_1, \theta_2).
\end{equation}
Our goal is to learn an embedding space without manually defined labels.

\subsection{Pseudo-labels by clustering}
In DeepCluster ~\cite{caron2018deep}, the authors utilize a signal provided by the convolutional structure of the random convnet, as a prior to the input signal. To bootstrap this signal, they need to use a large amount of training samples, e.g. ImageNet ~\cite{russakovsky2015imagenet} which contains 1.3M images uniformly distributed into 1000 classes. The goal of their work is to pre-train the model without labels.
However, in our work, we focus on the metric learning task and study the signal provided by pre-trained models, which allows us to learn an embedding space even with a few thousand samples.
Using the pre-trained network is a common practice in deep metric learning ~\cite{iscen2018mining, ye2019unsupervised}. The pre-trained convolutional neural network on ImageNet ~\cite{russakovsky2015imagenet} classification task can usually provide decent signal for image retrievals ~\cite{oh2016deep}.
Oh \etal ~\cite{oh2016deep} shows representations provided by Inception-V1 ~\cite{szegedy2015going} achieves reasonable performance on standard benchmarks of metric learning. 
The idea of this work is to exploit such pre-trained signal to bootstrap the metric learning process in an unsupervised manner.
For simplicity, we use k-means to cluster the feature vectors provided by the convnet and use the subsequent cluster assignments $y_i$
as pseudo-labels to initialize our proposed loss functions. 
After the initialization step, cluster reassignment is conducted on the embedding vectors instead of feature vectors.

\subsection{Multi-similarity Loss}
In general, our method is compatible with any metric learning loss function. Here we choose multi-similarity loss ~\cite{wang2019multi} because it shows the state of the art performance on supervised metric learning. However, we also show some experimental results using other popular metric learning loss functions in Section \ref{exp}. 

There are two steps in the multi-similarity method ~\cite{wang2019multi}. The first step is to mine hard sample pairs based on the cosine similarities between the corresponding embedding vectors.
The goal of sampling hard examples is to speed up the training process and extract informative pairs. 
Let $(x_1, y_1), (x_2, y_2), ..., (x_n, y_n)$ be the given instances, with $y_i$ as the pseudo-labels obtained from clustering assignments.
Let $x_i$ be the anchor. 
Define $(x_i, x_j)$ as a negative pair chosen as 
\begin{equation}
    S(x_i, x_j) > \operatorname*{min}_{y_h=y_i} S(x_i, x_h) - \epsilon.
\end{equation}
Similarly, $(x_i, x_k)$ is a positive pair chosen as 
\begin{equation}
    S(x_i, x_k) < \operatorname*{max}_{y_h \neq y_i} S(x_i, x_h) + \epsilon.
\end{equation}
$S$ represents the similarity between two examples. $\epsilon > 0$ controls the margin. 
Denote the set of chosen positive and negative pairs as $P_i$ and $N_i$ respectively. With the chosen training pairs, 
we can minimize the multi-similarity loss. The multi-similarity (MS) is defined as
\begin{multline}\label{eq:multi}
L_{MS} = \sum_{i=1}^{n} (\frac{1}{\alpha} \log(1 + \sum_{l \in P_i} e^{-\alpha(S_{il}-\lambda)}) + \\
    \frac{1}{\beta} \log(1 + \sum_{l \in N_i} e^{\beta(S_{il}-\lambda)}))
\end{multline}
We use the same $\alpha, \beta, \lambda$ and $\epsilon$ as in the original multi-similarity framework throughout our experiments.

\subsection{Predicting Image Rotations}
To overcome the challenge of unreliable clustering assignment, we aim to learn useful representations, independently of the quality of the clustering assignments.
To this end, we exploit recent advancements in self-supervised approaches for representation learning. 
For example, the network can be trained on a pretext task including colorizing grayscale images ~\cite{zhang2016colorful}, image inpainting ~\cite{pathak2016context}, image jigsaw puzzle ~\cite{noroozi2016unsupervised}, predicting image rotations ~\cite{gidaris2018unsupervised}.
We propose to add a self-supervised task to our metric learning loss. In particular, in this work, we focus on the pretext task proposed in ~\cite{gidaris2018unsupervised} where a model is trained to predict image rotations (0, 90, 180 and 270 degrees). They have shown state of the art performance on standard evaluation benchmarks in self-supervised learning.

Given a set of n images ${x_1, . . . , x_n}$, let $x_{ik}$ for $k=1, 2, 3, 4$ be the rotated version (0, 90, 180, 270 degrees) of $x_i$ and $z_{ik} = k$.  We learn
the parameters $\theta_1$ of the feature extraction backbone, $f_{\theta_1}$, jointly with parameters $\theta_3$ of a mapping, $h_{\theta_3}$, from features extractors to the predicted labels of rotation classification. Formally, we have
\begin{equation}\label{eq:rot}
L_{rot} = \frac{1}{n} \sum_{k=1}^{4}\sum_{i=1}^{n} L(h_{\theta_3}(f_{\theta_1}(x_{ik})), z_{ik})
\end{equation}
where $L$ is the cross-entropy loss.

\subsection{Proposed Loss Function}
Combining \eqref{eq:multi} and \eqref{eq:rot} gives us the following loss functions:
\begin{equation}\label{eq:loss}
L_{UDML-SS} = L_{MS}(\theta_1, \theta_2, x, y) + \eta L_{rot}(\theta_1, \theta_3, x, z)
\end{equation}
where $x$ is the training example, $y$ is pseudo-label given by clustering assignment and $z$ is the label for the pretext rotation task. $\eta > 0$ is a tuning parameter that balances the contributions of the metric learning loss and the rotation prediction loss. $\theta_1$ represents the parameter of feature extraction backbone, while $\theta_2$ and $\theta_3$ are the parameters of the metric learning layers and pretext task layers respectively. In other words, we use a single feature extraction network with two heads: one for metric learning and another for the pretext task.
Note that the input images for metric learning are not rotated because we believe that we should keep the self-supervised pretext task separated to avoid ``contamination'' from any unreliability in the cluster assignments. Empirically we also find that this choice achieves better performance. Given the loss function \ref{eq:loss}, UDML-SS iteratively learns and clusters the embedding vectors.

\section{Experimental Results}\label{exp}

\subsection{Datasets}
We conduct experiments on three standard datasets for metric learning: CUB-200-2011 ~\cite{wah2011caltech}, Cars-196 ~\cite{krause20133d} and Stanford Online Products ~\cite{oh2016deep}. We follow the same training/testing data split as ~\cite{oh2016deep}. 
\begin{itemize}
    \item The CUB-200-2011 (CUB) dataset has 11,788 images in total. We use the first 100 categories (5,864 images) for training and the remaining 100 categories for testing.
    \item The Cars-196 (Cars) dataset contains 16,185 images of 196 classes of cars. The first 98 model categories are used for training, and the rest for testing.
    \item  The Stanford Online Products dataset consists of 120,053 images of 22,634 online products from eBay.com. We use the first 11,318 products (59,551 images) for training and the remaining 11,316 products (60,502 images) for testing.
\end{itemize}

\begin{figure}[h!]
\begin{center}
\includegraphics[width=0.45\textwidth]{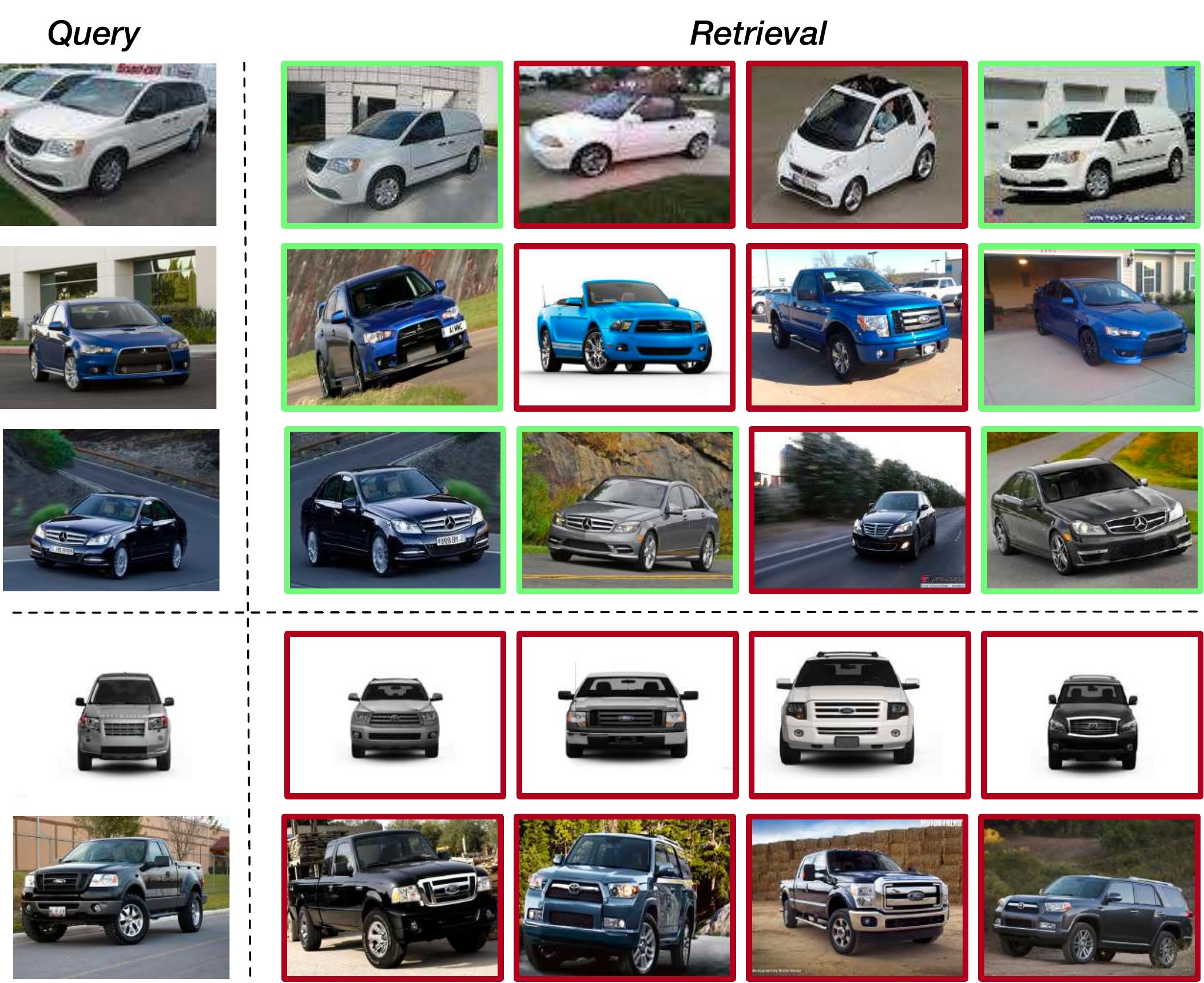}
\end{center}
  \caption{Retrieved images of sample queries on Cars196 dataset. The positive (negative) retrieved results are framed in green (red).  The last two rows show the failure cases. Best
viewed on a monitor zoomed in.}
\label{fig:car}
\end{figure}

\begin{figure*}[h!]
\begin{center}
\includegraphics[width=0.95\textwidth]{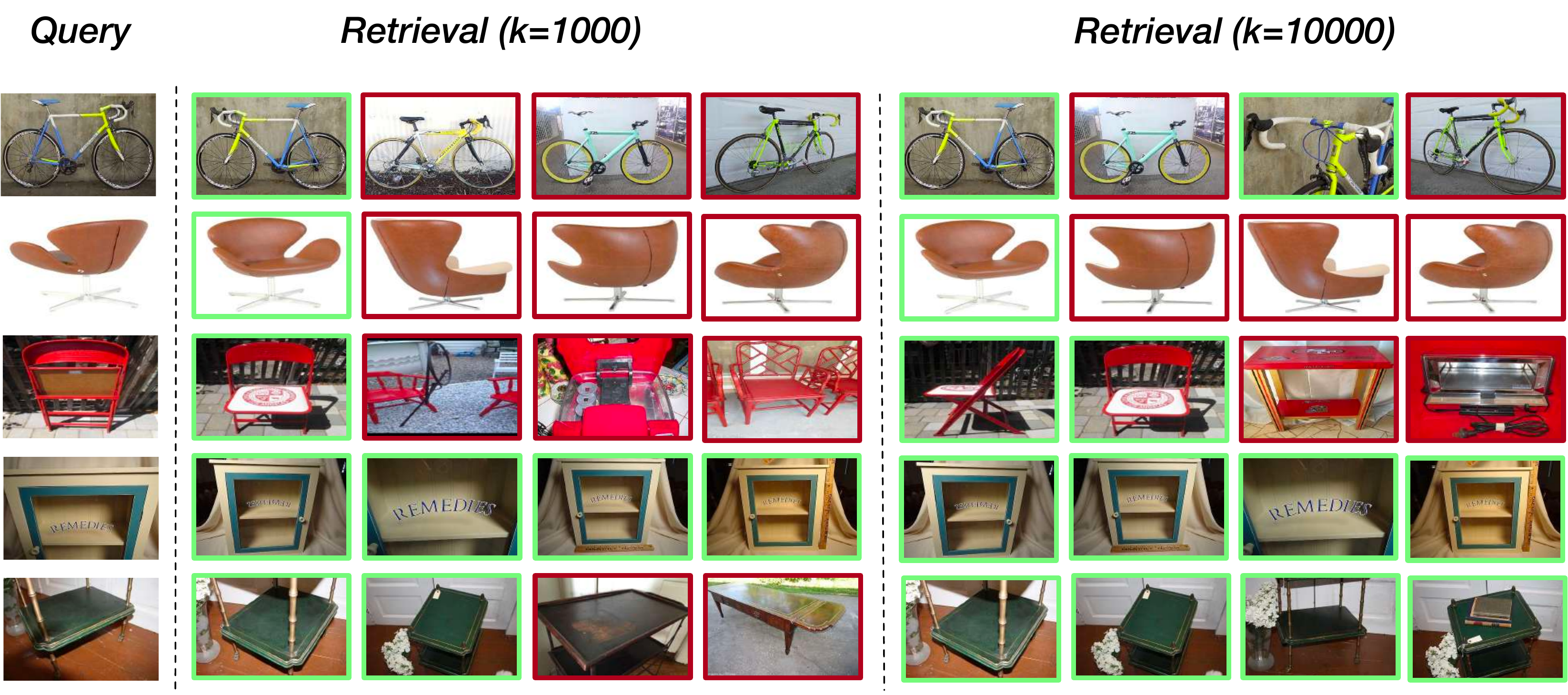}
\end{center}
  \caption{Retrieved images of sample queries on Stanford Online Products dataset for $k = 1000, 10000$. The positive (negative) retrieved results are framed in green (red). }
\label{fig:product}
\end{figure*}

\subsection{Experimental Settings}
Our method was implemented in PyTorch. We utilize
Inception-V1 ~\cite{szegedy2015going}  pre-trained on ImageNet ~\cite{russakovsky2015imagenet} as the backbone network, and
fine-tuned it for our task. We also show experimental results on some other network architectures in Section \ref{sec:backbone}.
We add two separate fully connected layers (512-dim) on the top of the network following the global pooling layer. The first one is for the embedding layer and the second one is for rotation classification. All the input
images were cropped to 227 $\times$ 227. During the training phase, we use random cropping with random horizontal mirroring for data augmentation. In the testing phase, a single center-cropped image is the input for fine-grained retrieval as in ~\cite{wang2019multi}.
We use Adam ~\cite{kingma2014adam} optimizer for all experiments

For each mini-batch, we follow the sampling strategy used in ~\cite{wang2019multi} and we randomly choose a certain
number of classes, and then sample M = 5 examples
from each class for all datasets in our experiments. To implement the rotation loss, we rotate 16 images in the batch in all four considered directions (0, 90, 180, and 270 degrees) and get a batch size of 64 (16 unique images) ~\cite{gidaris2018unsupervised}. For CUB and Cars datasets, we use 100 clusters (i.e. $k=100$) while we show evaluations of Stanford Online Products with $k=1000, 10000$, as it contains a much larger number of samples. As shown in ~\cite{ye2019unsupervised}, the pre-trained Inception-V1
performs better on the CUB and Products dataset than on the Cars dataset.
Based on this fact, we set $\eta=0.5$ for Cars so that the contribution of self-supervision is increased to reduce the dependency of learned representations on the quality of the cluster assignments, while for CUB and Products, we set $\eta=0.1$. For more details on the performance of different $\eta$, see the appendix.
For all other hyperparameters, we use
the values provided in ~\cite{wang2019multi}. The similarity is measured with cosine similarity.

We evaluate our method on image retrieval task
by using the standard performance metric: Recall@K. Given a query image from the testing set, Recall@K is the probability that any correct matching occurs in the top-k retrieved images. We also provide Normalized Mutual Information (NMI)
to measure the clustering performance of the testing dataset.
NMI is defined by the ratio of the mutual information of clusters and ground truth labels to the arithmetic mean of their entropy ~\cite{schutze2008introduction}. 

\begin{table}
\begin{center}
\resizebox{0.48\textwidth}{!}{
\begin{tabular}{lllllll}
\hline
Method & R@1 & R@2 & R@4 & R@8 & NMI  \\
\hline
Supervised\\
\hline 
Lifted  ~\cite{oh2016deep}      & 46.9 & 59.8 & 71.2 & 81.5 & 56.4 \\
Angular  ~\cite{wang2017deep}     & 53.6 & 65.0 & 75.3 & 83.7 & 61.0 \\
Triplet   ~\cite{weinberger2009distance}    & 35.9 & 47.7 & 59.1 & 70.0 & 49.8 \\
Triplet hard  ~\cite{schroff2015facenet} & 40.6 & 52.3 & 64.2 & 75.0 & 53.4 \\
Multi-Sim ~\cite{wang2019multi} & 65.7 & 77.0 & 86.3 & 91.2 & -\\
\hline
Unsupervised\\
\hline 
Exemplar ~\cite{dosovitskiy2015discriminative}   & 38.2 & 50.3 & 62.8 & 75.0 & 45.0 \\
NCE  ~\cite{wu2018unsupervised}      & 39.2 & 51.4 & 63.7 & 75.8 & 45.1 \\
DeepCluster ~\cite{caron2018deep} & 42.9 & 54.1 & 65.6 & 76.2 & 53.0 \\
Rot-Only ~\cite{gidaris2018unsupervised} & 42.5 & 55.8 & 68.6 & 79.4 & 49.1\\ 
MOM   ~\cite{iscen2018mining}     & 45.3 & 57.8 & 68.6 & 78.4 & 55.0 \\
Instance  ~\cite{ye2019unsupervised}      & \textcolor{red}{46.2} & \textcolor{red}{59.0} & \textcolor{red}{70.1} & \textcolor{red}{80.2} & \textcolor{red}{55.4} \\
\hline
UDML-SS & \textbf{54.7}     & \textbf{66.9}     & \textbf{77.4}     & \textbf{86.1} & \textbf{61.4} \\
\hline
\end{tabular}
}
\end{center}
\caption{Experimental results (\%) on the CUB-200-2011 dataset in comparison with other methods.}
\label{tab:cub-main}
\end{table}

\subsection{Comparison with State-of-the-Art}
Tables \ref{tab:cub-main}, \ref{tab:car-main}, and \ref{tab:product-main} show quantitative results on the
CUB-200-2011, Cars196, and Stanford Online Products
datasets, respectively. MOM ~\cite{iscen2018mining} and Instance ~\cite{ye2019unsupervised} are two most recent state of art methods designed for unsupervised metric learning and they also utilize Inception-V1 ~\cite{szegedy2015going} as their backbone. The performance of other state-of-the-art unsupervised methods (Exemplar ~\cite{dosovitskiy2015discriminative}, NCE ~\cite{wu2018unsupervised}, Deep Cluster ~\cite{caron2018deep} and Prediction Image Rotation ~\cite{gidaris2018unsupervised}) on these three datasets ~\cite{ye2019unsupervised} are also listed in these tables.  For a fair comparison, we list evaluations using the same number of clusters ($k$) for UDML-SS and DeepCluster ~\cite{caron2018deep} methods. 
We also provide the performance of several popular supervised metric learning methods (Triplet ~\cite{weinberger2009distance}, Triplet Hard ~\cite{schroff2015facenet}, Lifted Structure ~\cite{oh2016deep}, Angular ~\cite{wang2017deep}, Multi-Similarity ~\cite{wang2019multi}) to show the relative performance of unsupervised metric learning. Note that only the multi-similarity work uses Inception-V2 ~\cite{ioffe2015batch}.

As shown in Table \ref{tab:cub-main}, UDML-SS outperforms all competing methods with a large
margin on CUB-200-2011 dataset. For example, we have achieved an 11.8\% and 8.5\% increase of Recall@1 compared to DeepCluster ~\cite{caron2018deep} and the instance method ~\cite{ye2019unsupervised}, respectively. Some qualitative results on CUB dataset are shown in the appendix. In Table \ref{tab:car-main} and \ref{tab:product-main}, we show the results on Cars196 and Product datasets. We observe UDML-SS achieves very competitive performance and outperforms all competing methods with a clear margin on these two datasets. 
In particular, UDML-SS achieves a 14.6\% boost of Recall@1 for Product dataset ($k=10000$) and a 3.8\% boost of Recall@1 on Cars196 dataset. 
When $k=1000$ for Product dataset, which is much smaller than the number of classes (11318), our method still outperforms the competing method ~\cite{ye2019unsupervised}.

It is also noteworthy that the performance of UDML-SS on CUB and Product datasets is much closer than expected to some supervised methods. However, there is still a large gap between supervised and unsupervised method on Car. Figure \ref{fig:car} shows some example queries and nearest neighbors on Car with both successful and failure examples using UDML-SS. 
Some failure examples showed in Figure \ref{fig:car} look very similar to the query except logos, which is the common feature that people use to discriminate different cars. 
In general, it is hard for the unsupervised method to detect these kinds of fine-grained differences and we leave it for future work.
In Figure \ref{fig:product}, we show a comparison of retrievals using different number of clusters on Product.

\begin{table}
\begin{center}
\resizebox{0.48\textwidth}{!}{
\begin{tabular}{lllllll}
\hline
Method & R@1 & R@2 & R@4 & R@8 & NMI  \\
\hline
Supervised\\
\hline 
Lifted  ~\cite{oh2016deep}      & 59.9 & 70.4 & 79.6 & 87.0 & 57.8\\
Angular  ~\cite{wang2017deep}     & 71.3 & 80.7 & 87.0 & 91.8 & 62.4 \\
Triplet    ~\cite{weinberger2009distance}   & 45.1 & 57.4 & 69.7 & 79.2 & 52.9 \\
Triplet\_hard ~\cite{schroff2015facenet} & 53.2 & 65.4 & 74.3 & 83.6 & 55.7 \\
Multi-Sim ~\cite{wang2019multi} & 84.1 & 90.4 & 94.0 & 96.5  & -\\

\hline 
Unsupervised\\
\hline
Exemplar ~\cite{dosovitskiy2015discriminative}  & 36.5 & 48.1 & 59.2 & 71.0 & 35.4   \\
NCE  ~\cite{wu2018unsupervised} & 37.5 & 48.7 & 59.8 & 71.5 & 35.6        \\
DeepCluster ~\cite{caron2018deep} & 32.6 & 43.8 &57.0 &69.5 & 38.5 \\
Rot-Only ~\cite{gidaris2018unsupervised} & 33.3 & 44.6 & 56.4 & 68.5 & 32.7\\ 
MOM  ~\cite{iscen2018mining} & 35.5 &48.2 &60.6 &72.4 & \textcolor{red}{38.6}        \\
Instance ~\cite{ye2019unsupervised} & \textcolor{red}{41.3}& \textcolor{red}{52.3}& \textcolor{red}{63.6}& \textcolor{red}{74.9} & 35.8    \\
\hline
UDML-SS & \textbf{45.1}     & \textbf{56.1}     & \textbf{66.5}     & \textbf{75.7} & 34.4 \\
\hline
\end{tabular}
}
\end{center}
\caption{Experimental results (\%) on the Cars196 dataset in comparison with other methods.}
\label{tab:car-main}
\end{table}

\begin{table}
\begin{center}
\resizebox{0.48\textwidth}{!}{
\begin{tabular}{lllllll}
\hline
Method & R@1 & R@10 & R@100 & NMI  \\
\hline 
Supervised \\
\hline 
Lifted ~\cite{oh2016deep}& 62.6 & 80.9 & 91.2 & 87.2 \\
Angular ~\cite{wang2017deep}& 67.9 & 83.2 & 92.2 & 87.8 \\
Triplet ~\cite{weinberger2009distance}& 53.9 & 72.1 & 85.7 & 86.3 \\
Triplet hard ~\cite{schroff2015facenet}&  57.8 & 75.3 & 88.1 & 86.7 \\
Multi-Sim ~\cite{wang2019multi} & 78.2 & 90.5 & 96.0& -\\
\hline 
Unsupervised \\
\hline 
Exemplar ~\cite{dosovitskiy2015discriminative}   & 45.0 & 60.3 & 75.2 & 85.0 \\
NCE   ~\cite{wu2018unsupervised}      & 46.6 & 62.3 & 76.8 & 85.8 \\
DeepCluster ~\cite{caron2018deep} & 46.1 & 61.1 & 76.0 & 85.3\\
Rot-Only ~\cite{gidaris2018unsupervised} & 40.1 & 54.6 & 70.1 & 82.7\\
MOM   ~\cite{iscen2018mining}      & 43.3 & 57.2 & 73.2 & 84.4 \\
Instance     ~\cite{ye2019unsupervised}   & \textcolor{red}{48.9} & \textcolor{red}{64.0} & \textcolor{red}{78.0} & \textcolor{red}{86.0} \\
\hline
UDML-SS (k=1000) & \textbf{54.4}     & \textbf{70.0}     & \textbf{82.9}     & \textbf{86.5} \\
UDML-SS (k=10000) & \textbf{63.5}     & \textbf{78.0}     & \textbf{88.6}     & \textbf{88.4} \\
\hline
\end{tabular}
}
\end{center}
\caption{Experimental results (\%) on the Product dataset in comparison with other methods.}
\label{tab:product-main}
\end{table}

\begin{table}
\begin{center}
\resizebox{0.48\textwidth}{!}{
\begin{tabular}{lllllll}
\hline
Method & R@1 & R@2 & R@4 & R@8 & NMI  \\
\hline
CUB\\
\hline
Instance ~\cite{ye2019unsupervised}       & 46.2 & 59.0 & 70.1 & 80.2 & 55.4 \\
Ours \scriptsize{($\eta = 0$)}     & 51.7 & 63.7 & 74.6 & 84.2 & 59.0 \\
Ours &  \textbf{54.7}     & \textbf{66.9}     & \textbf{77.4}     & \textbf{86.1} & \textbf{61.4}\\
\hline
Cars\\
\hline 
Instance ~\cite{ye2019unsupervised}       & 41.3 & 52.3    & 63.6 &    74.9 &    \textbf{35.8} \\
\hline 
Ours \scriptsize{($\eta = 0$)}   &  38.1 &    48.2 &    58.7 &    69.1 &    32.4 \\
Ours & \textbf{45.1}     & \textbf{56.1}     & \textbf{66.5}     & \textbf{75.7} & 34.4 \\
\hline 
\end{tabular}
}
\resizebox{0.48\textwidth}{!}{
\begin{tabular}{llllll}
\hline 
Method & R@1 & R@100 & R@1000 & NMI  \\
\hline
Product\\
\hline
Instance  ~\cite{ye2019unsupervised}   & 48.9 & 64.0 & 78.0 & 86.0 \\
\hline 
Ours \tiny{($\eta = 0$, $k=1000$)} & 53.2     & 68.6    & 82.0    & 86.3 \\
Ours \tiny{($k=1000$)}& \textbf{54.4}     & \textbf{70.0}     & \textbf{82.9}     & \textbf{86.5} \\
\hline
Ours \tiny{($\eta = 0$, $k=10000$)} & 63.4 & 77.4 & 87.6 & \textbf{88.4} \\
Ours \tiny{($k=10000$)}& \textbf{63.5}    & \textbf{78.0} &    \textbf{88.6} &    \textbf{88.4} \\
\hline
\end{tabular}
}
\end{center}
\caption{Ablation results (\%) on the CUB-200-2011, Cars196 and Stanford Online Products datasets in comparison. The first row for each dataset is from the state of art unsupervised metric learning method ~\cite{ye2019unsupervised} while the second and the third row is the performance of UDML-SS without rotation loss, i.e. $\eta=0$ and with rotation loss.}
\label{tab:all-ablation}
\end{table}

\subsection{Ablation Study}
We also conduct ablation study of the
proposed unsupervised framework. Figure \ref{tab:all-ablation} shows the performance of UDML-SS with and without self-supervision loss. For all three datasets, metric learning loss equipped with k-means already achieves a decent performance. With additional rotation loss, the performance is further improved with a clear margin. For the Car dataset, the performance difference is up to 7\% of Recall@1 between our method with and without rotation loss, which shows the importance of the rotation loss to UDML-SS. 
It is surprising that UDML-SS without the rotation loss still achieves better results than ~\cite{ye2019unsupervised} on CUB and Product datasets. We speculate it is because the pre-trained Inception-V1 ~\cite{szegedy2015going} itself provides a stronger signal to CUB and Product datasets than to the Car dataset. For more details about the performance of pre-trained Inception-V1 network, see ~\cite{oh2016deep}.


\subsection{On Different Metric Learning Losses}
\begin{figure}[h!]
\begin{center}
\includegraphics[width=0.5\textwidth]{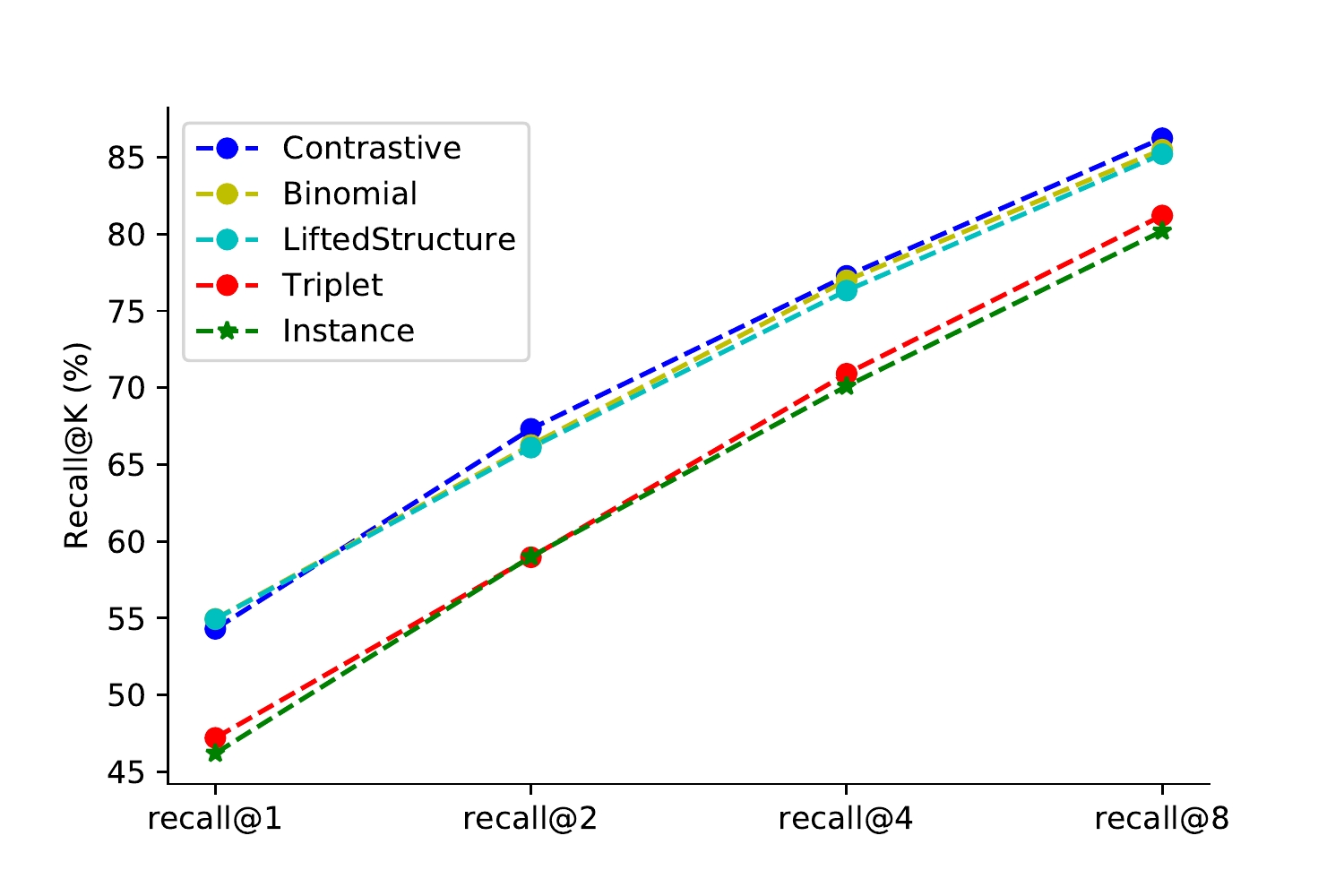}
\end{center}
  \caption{This Figure shows Recall@K of our method on CUB-200-2011 with different metric learning losses. For comparison, we also add Instance ~\cite{ye2019unsupervised} to the figure.}
\label{fig:cub-extra}
\end{figure}

In this section, we study the performance of UDML-SS using different metric learning loss functions. We conduct all the following experiments on the CUB dataset.
It is interesting to see that in Figure \ref{fig:cub-extra} performances for Contrastive Loss, Binomial Loss and Lifted Structure Loss are very similar and all of them outperform the competing methods. However, Triplet Loss ~\cite{schroff2015facenet} does not work well, possibly because the success of triplet loss depends more on the label correctness. 

\subsection{On the Embedding Size}
\begin{figure}[h!]
\begin{center}
\includegraphics[width=0.5\textwidth]{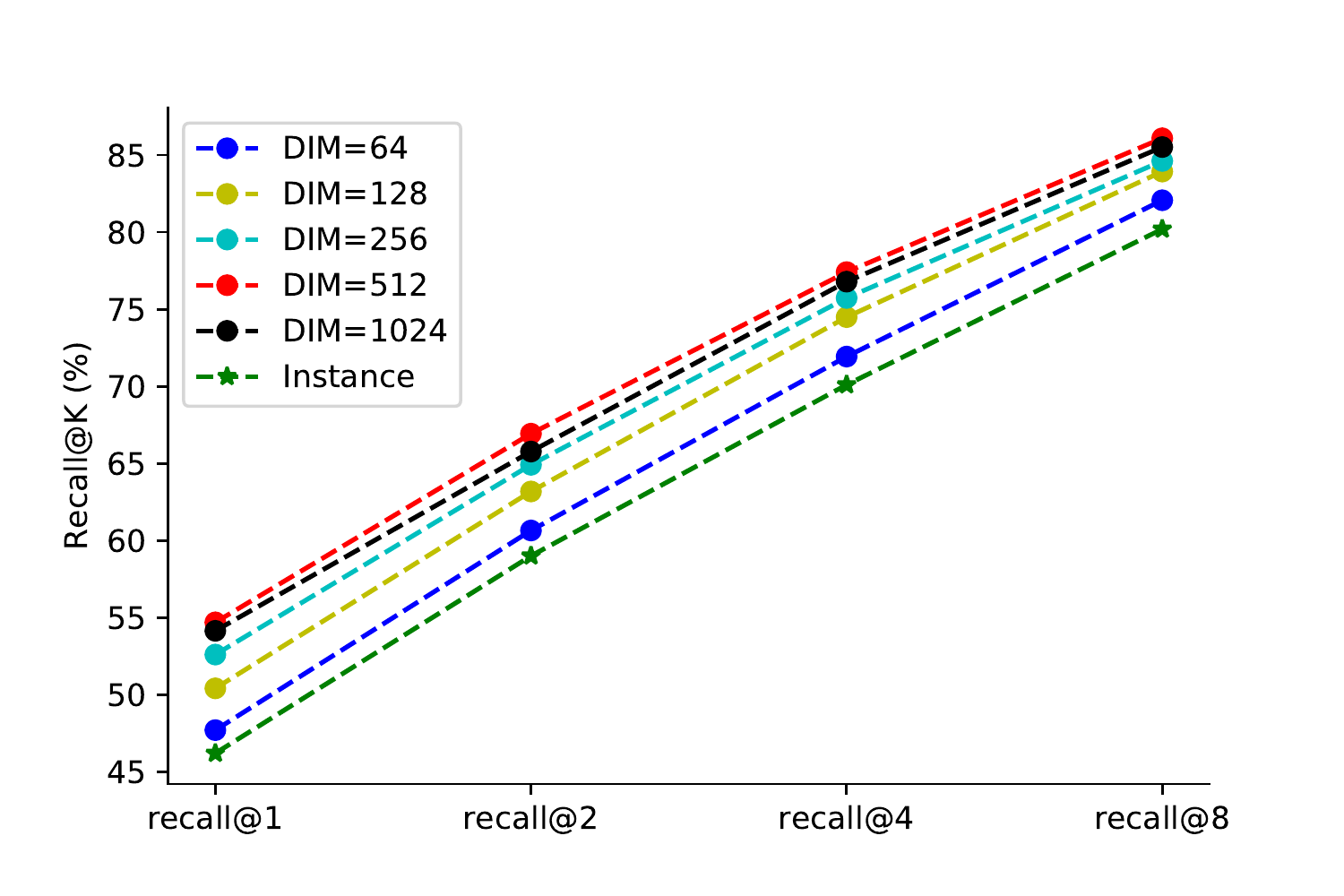}
\end{center}
  \caption{This Figure shows Recall@K of our method on CUB-200-2011 with different embedding vector size. The Instance ~\cite{ye2019unsupervised} method is listed for comparison. }
\label{fig:cub-embedding}
\end{figure}

Following ~\cite{schroff2015facenet}, we study the performance of our proposed loss with different embedding sizes of \{64, 128, 256, 512, 1024\}. As shown in Figure \ref{fig:cub-embedding}, the performance is increased consistently with the embedding dimension except at 1024 on the CUB dataset. Our method with embedding dimension at 512 and 1024 performs similarly. We observe a similar pattern in the evaluations of Car and Product datasets. These results are provided in the appendix.


\subsection{Choosing the Number of Clusters}
\begin{figure}[h!]
\begin{center}
\includegraphics[width=0.5\textwidth]{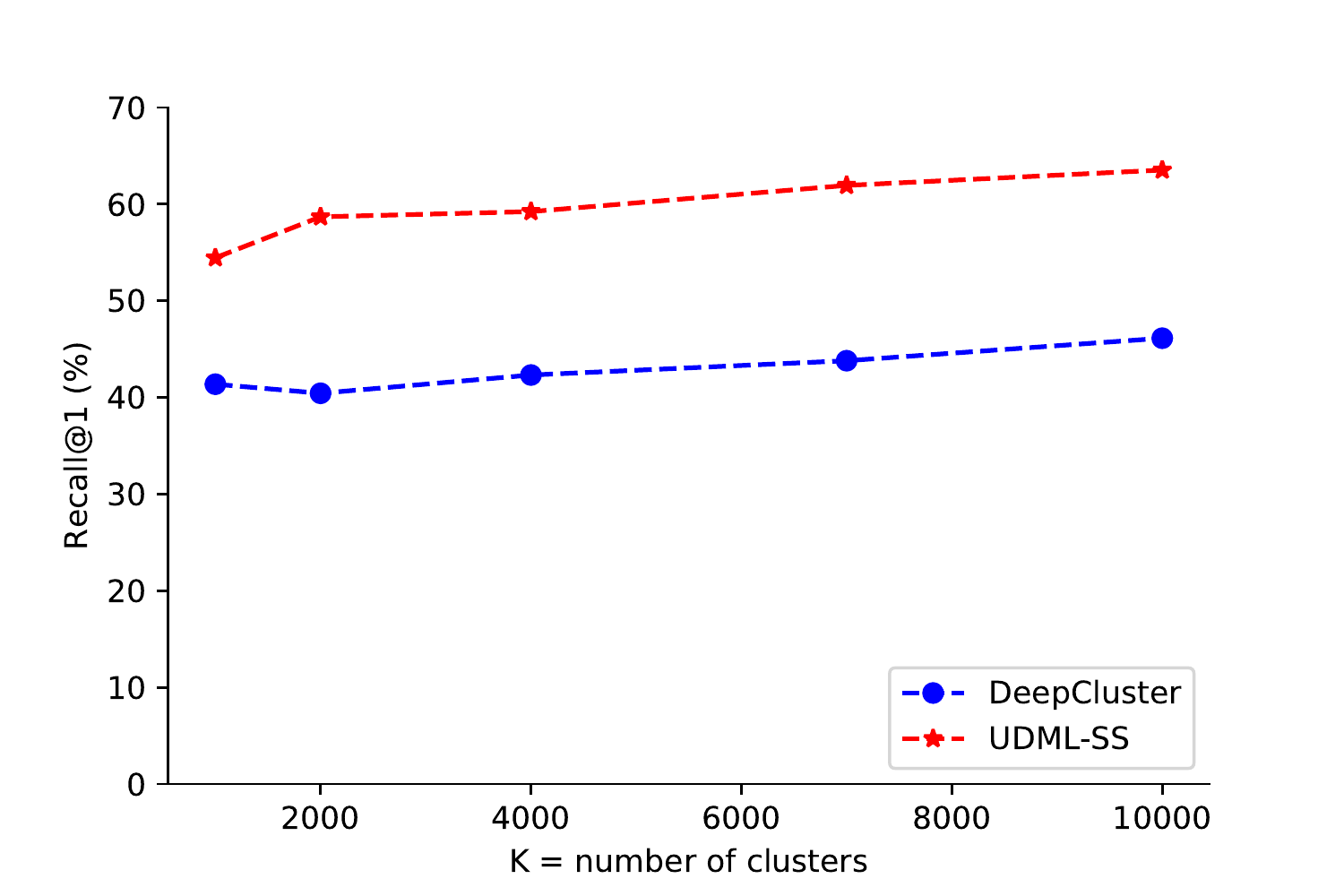}
\end{center}
  \caption{This figure shows Recall@1 of our method compared with DeepCluster on Product for various choices of $k$.}
\label{fig:product_clustervs}
\end{figure}

In this section, we measure the impact of the number of
clusters ($k$) used in k-means on the performance of different datasets. In Table \ref{tab:cub-cluster}, we show the results using 50, 100, 250, 500, 1000 clusters. From the table for CUB dataset, we can see from a wide range of number of clusters (e.g. 50, 100, 250, 500), our method can achieve a better result than the competing method ~\cite{ye2019unsupervised}. However, when the cluster size gets too large (e.g. k=1000), our method's performance drops below that of the competing method. UDML-SS using different clusters on Car dataset shows similar performance trend as CUB dataset (see the appendix for details).
For Product dataset, we show the performance of UDML-SS compared with DeepCluster ~\cite{caron2018deep} using the different number of clusters in Figure \ref{fig:product_clustervs}. 
\begin{table}
\begin{center}
\resizebox{0.48\textwidth}{!}{
\begin{tabular}{llllll}
\hline 
k     & R@1 & R@2 & R@4 & R@8 & NMI  \\
\hline 
Instance ~\cite{ye2019unsupervised}       & 46.2 & 59.0 & 70.1 & 80.2 & 55.4 \\
\hline 
50   & 50.7     & 63.0      & 74.4  & 84.1  & 57.0 \\
100  & \textbf{54.7} & \textbf{66.9} & \textbf{77.4} & \textbf{86.1} & \textbf{61.4} \\
250 & 52.1 & 64.5 & 75.4 & 83.9 & 56.6 \\
500  & 49.4 & 61.4 & 73.1 & 83.3 & 56.5 \\
1000 & 46.2     &58.1    & 70.4    & 81.5    & 52.3 \\
\hline
\end{tabular}
}
\end{center}
\caption{This table shows evaluations of our method on CUB-200-2011 for various choices of $k$.}
\label{tab:cub-cluster}
\end{table}

\subsection{On Different ConvNet Backbones} \label{sec:backbone}

\begin{table}
\begin{center}
\resizebox{0.48\textwidth}{!}{
\begin{tabular}{lllllll}
\hline
Method & R@1 & R@2 & R@4 & R@8 & NMI  \\
\hline
Inception-V1 ~\cite{szegedy2015going} & 54.7   & 66.9    & 77.4    & 86.1 & 61.4\\
Inception-V2 ~\cite{ioffe2015batch} & \textbf{63.7} & \textbf{75.0} & \textbf{83.8} & \textbf{90.1} & \textbf{67.1} \\
ResNet34  ~\cite{he2016deep} & 59.0 & 70.6 & 80.4 & 88.1 & 63.4 \\
ResNet50  ~\cite{he2016deep} & 60.1 & 71.6 & 81.8 & 88.5 & 64.5 \\
ResNet101 ~\cite{he2016deep} & 61.8 & 73.0 & 82.6 & 89.6 & 65.5 \\
\hline
\end{tabular}
}
\end{center}
\caption{Experimental results (\%) on the CUB-200-2011 dataset with different backbones.}
\label{tab:cub-backbone}
\end{table}
Here we show the performance of UDML-SS using different feature extractions backbones. The experiment is conducted on CUB dataset. From Table \ref{tab:cub-backbone}, we observe that Inception-V2 ~\cite{ioffe2015batch} and ResNet ~\cite{he2016deep} can improve the performance by a large margin, while the performance difference between different ResNet is relatively small. 
It is noteworthy that the performance of UDML-SS with Inception-V2 is very close to that of supervised multi-similarity method ~\cite{wang2019multi}, which also uses Inception-V2.

\subsection{Learning from Scratch}
So far, we showed that UDML-SS can perform very well with a chosen pre-trained network, e.g. Inception-V1 ~\cite{szegedy2015going}. Now following \cite{ye2019unsupervised}, we evaluate the performance of UDML-SS using ResNet18 ~\cite{he2016deep} without pre-training. Table \ref{tab:product-scratch} shows the performance of UDML-SS on Product dataset. The competing methods' performance are originally from ~\cite{ye2019unsupervised}. It is impressive that our method with random-initialized network can still outperform all other methods with a very large margin.

\begin{table}
\begin{center}
\resizebox{0.48\textwidth}{!}{
\begin{tabular}{lllllll}
\hline
Methods  & R@1  & R@10 & R@100 & NMI  \\
\hline 
Random   & 18.4 & 29.4 & 46.0  & 79.8 \\
Exemplar ~\cite{dosovitskiy2015discriminative} & 31.5 & 46.7 & 64.2  & 82.9 \\
NCE  ~\cite{wu2018unsupervised}    & 34.4 & 49.0 & 65.2  & 84.1 \\
MOM   ~\cite{iscen2018mining}   & 16.3 & 27.6 & 44.5  & 80.6 \\
Instance ~\cite{ye2019unsupervised} & \textcolor{red}{39.7} & \textcolor{red}{54.9} & \textcolor{red}{71.0}  & \textcolor{red}{84.7} \\
\hline 
UDML-SS  & \textbf{59.2}    & \textbf{73.9}    & \textbf{85.1}     &   \textbf{87.6}  \\
\hline
\end{tabular}
}
\end{center}
\caption{Experimental results (\%) on the Product dataset with random-initialized network.}
\label{tab:product-scratch}
\end{table}


\section{Conclusion}
In this work, we have presented a new unsupervised metric learning framework (UDML-SS), which for the first time, combines clustering, self-supervised learning, and metric learning. 
In particular, we iteratively cluster embedding vectors using k-means and update embedding vectors by optimizing a multi-task loss function, which considers both similarity learning task and image rotation prediction task. 
We have demonstrated the effectiveness of the
proposed framework on three popular benchmark datasets in metric learning. UDML-SS obtains a new state-of-the-art performance on these datasets.
In addition, we explore the performance of UDML-SS with various popular metric learning loss functions. We empirically show that UDML-SS obtains state of the art performance even with a randomly initialized network.

{\small
\bibliographystyle{ieee_fullname}
\bibliography{egbib}
}

\section{Appendix}
In this appendix, we show more detailed experimental results to support our paper. 
We add more details of experiments on three standard datasets for metric learning: CUB-200-2011 ~\cite{wah2011caltech}, Cars-196 ~\cite{krause20133d} and Stanford Online Products ~\cite{oh2016deep}. 
\begin{figure}[h!]
\begin{center}
\includegraphics[width=0.4\textwidth]{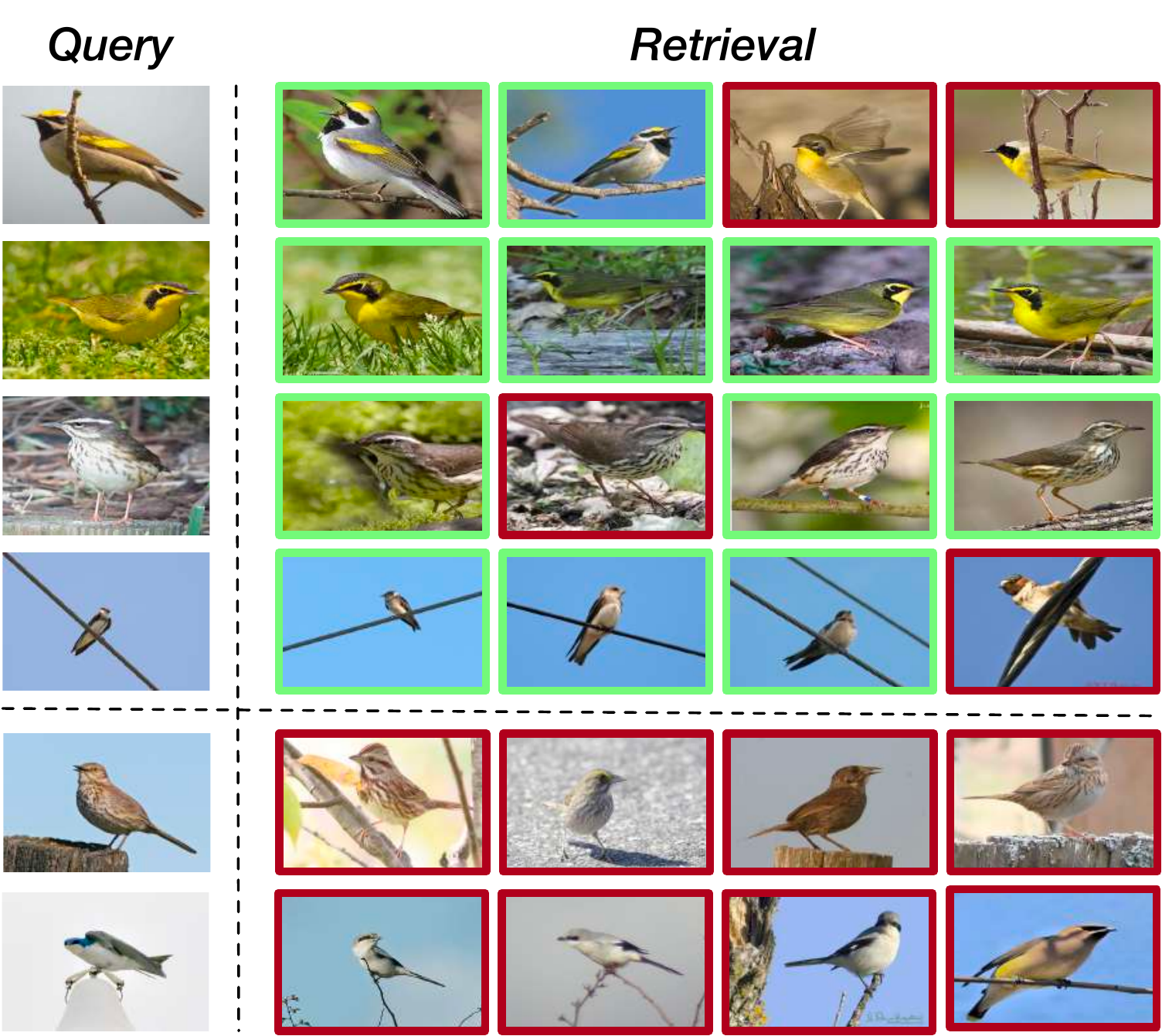}
\end{center}
  \caption{Retrieved images of sample queries on CUB-200-2011 dataset. The positive (negative) retrieved results are framed in green (red). The last two rows show the failure cases. Best viewed on a monitor zoomed in.}
\label{fig:cub}
\end{figure}

\subsection{On the Embedding Size}
In Table \ref{tab:car-embedding} and  \ref{tab:product-embedding}, we show the experimental results of UDML-SS on Car and Product datasets with different embedding sizes. 
\begin{table}[h!]
\begin{center}
\begin{tabular}{llllll}
\hline 
Embedding Size      & R@1 & R@2 & R@4 & R@8  \\
\hline 
64   & 36.0 & 45.6 & 56.7 & 67.4 \\
128  & 41.2 & 50.8 & 61.6 & 72.0 \\
256  & 43.2 & 53.4 & 63.7 & 73.8  \\
512  & 45.1 & 56.1 & \textbf{66.5} & \textbf{75.7} \\
1024 & \textbf{46.1} & \textbf{56.2} & 66.4 & 75.6 \\
\hline
\end{tabular}
\end{center}
\caption{Recall@K (\%) of UDML-SS on Car dataset with different embedding vector size. }
\label{tab:car-embedding}
\end{table}

\begin{table}[h!]
\begin{center}
\begin{tabular}{llllll}
\hline 
Embedding Size      & R@1 & R@2 & R@100 \\
\hline 
64   &   62.5    &   77.2    &    88.2     \\
128   &   62.9    &   77.7   &     88.4     \\
256    &  63.3    &   77.9    &    \textbf{88.6}  \\
512  &     \textbf{63.5}  &     \textbf{78.0}      &  \textbf{88.6}  \\
1024  &    \textbf{63.5}   &    77.9     &   88.5  \\
\hline
\end{tabular}
\end{center}
\caption{Recall@K (\%) of UDML-SS on Product dataset with different embedding vector size. }
\label{tab:product-embedding}
\end{table}

\subsection{Choosing the Number of Clusters}
\begin{figure}[h!]
\begin{center}
\includegraphics[width=0.5\textwidth]{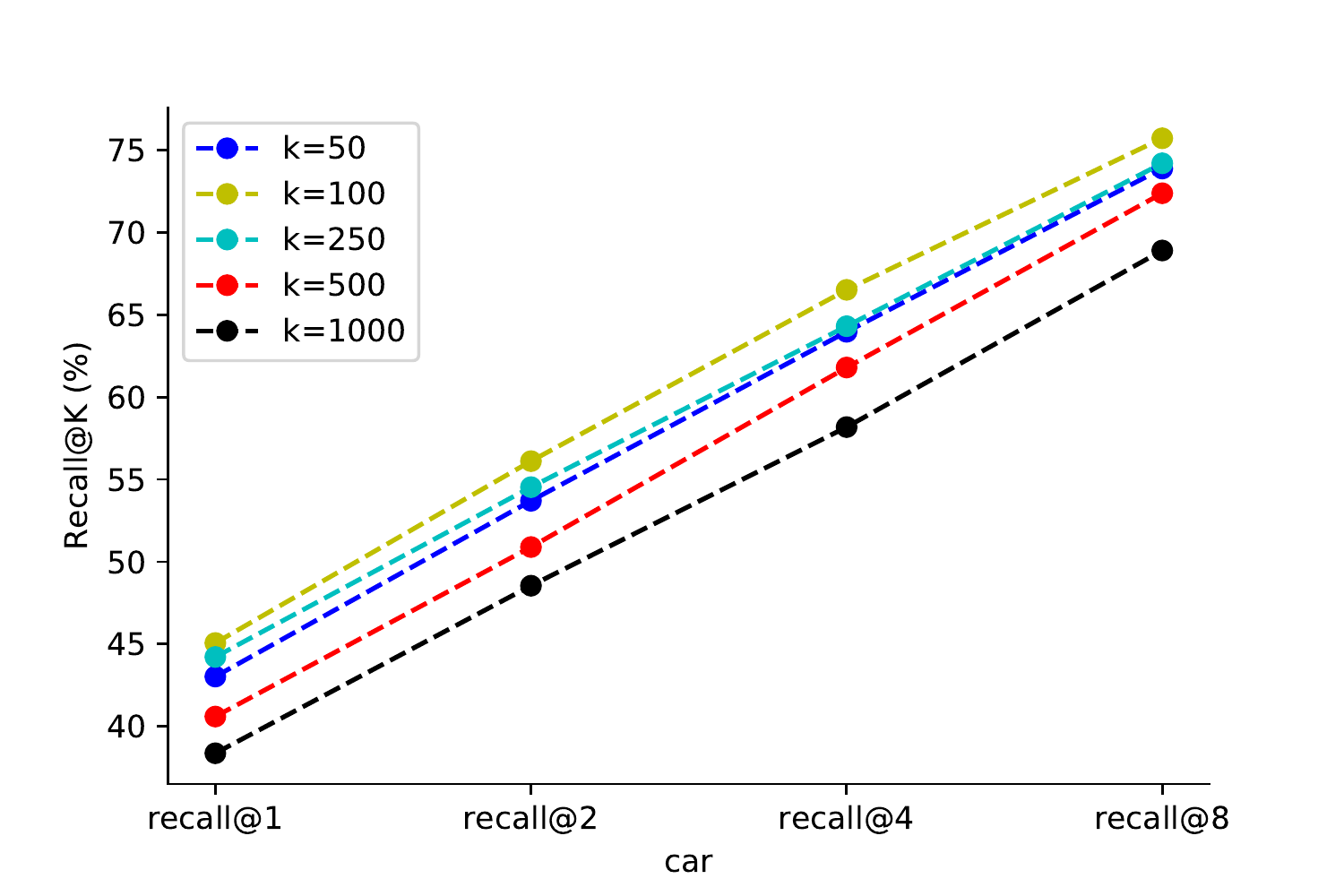}
\end{center}
  \caption{ Performance under different $k$ i.e. number of clusters of UDML-SS on Car dataset}
\label{fig:car_cluster}
\end{figure}

\begin{figure}[h!]
\begin{center}
\includegraphics[width=0.5\textwidth]{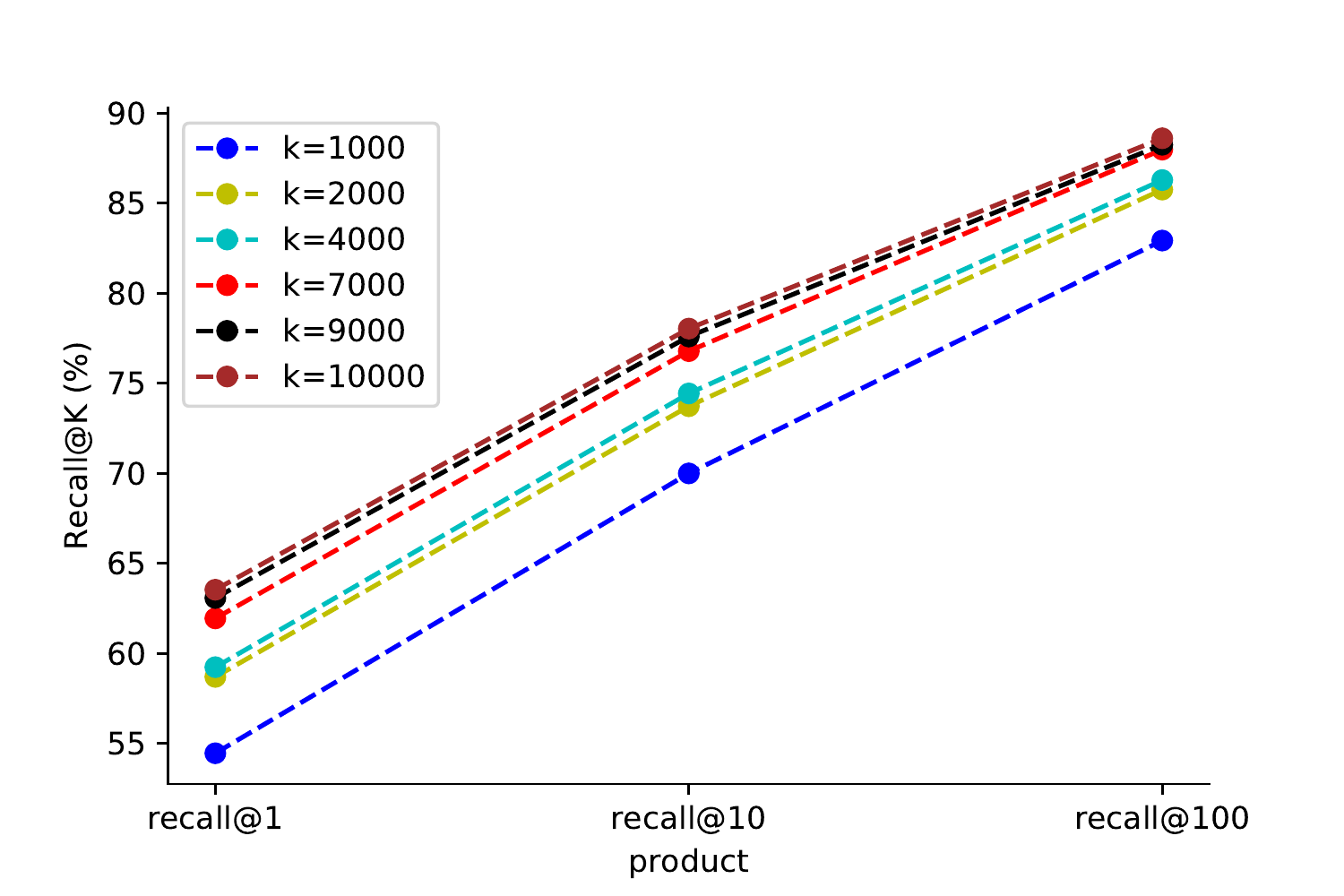}
\end{center}
  \caption{ Performance under different $k$ i.e. number of clusters of UDML-SS on Product dataset}
\label{fig:product_cluster}
\end{figure}

In Figure \ref{fig:car_cluster} and \ref{fig:product_cluster}, we show how $k$, i.e. the number of clusters affect our evaluations. 

\subsection{How $\eta$ affects evaluation results}
\begin{figure}[h!]
\begin{center}
\includegraphics[width=0.5\textwidth]{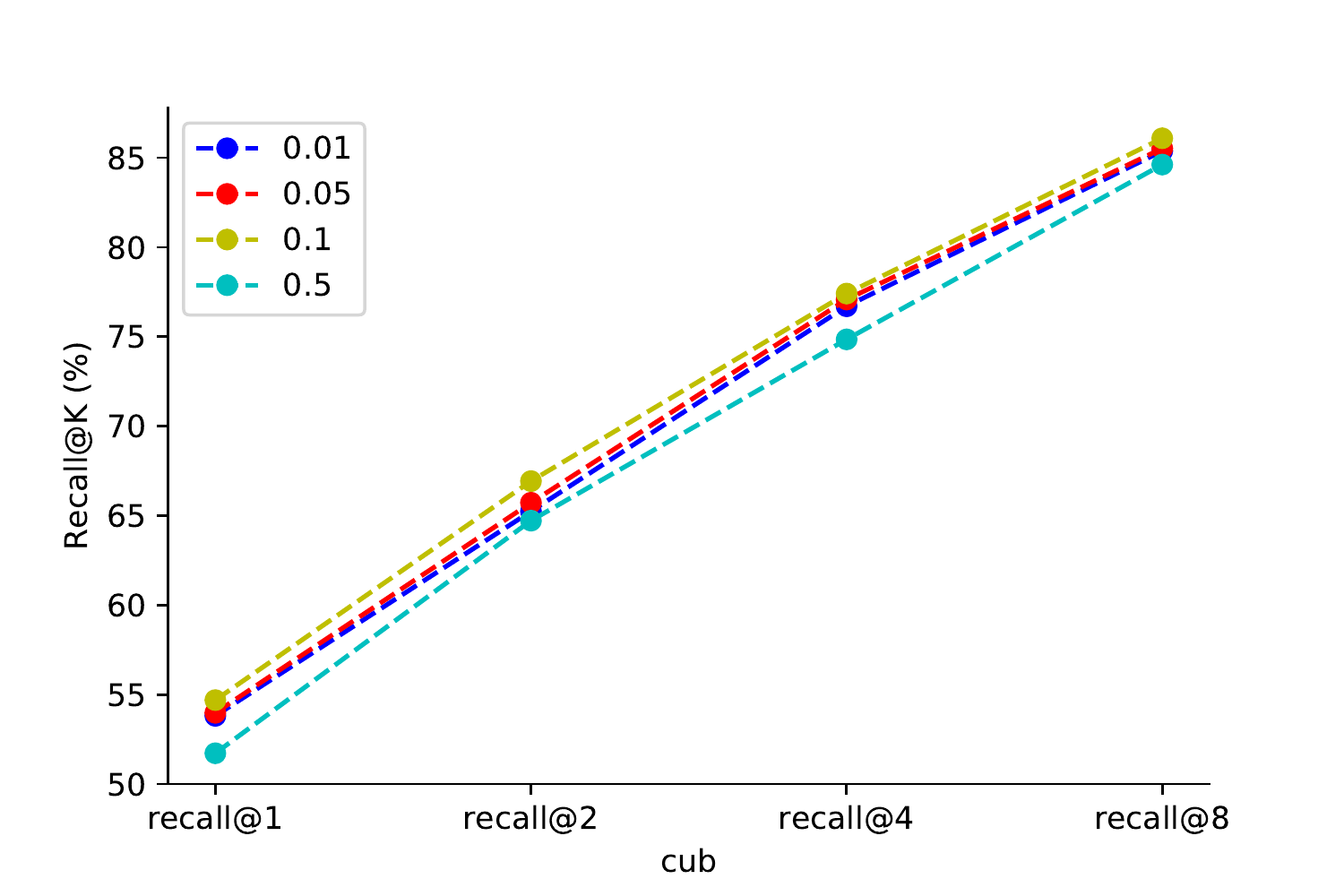}
\end{center}
  \caption{ Performance under different $\eta$ of UDML-SS on CUB dataset}
\label{fig:cub_etas}
\end{figure}

\begin{figure}[h!]
\begin{center}
\includegraphics[width=0.5\textwidth]{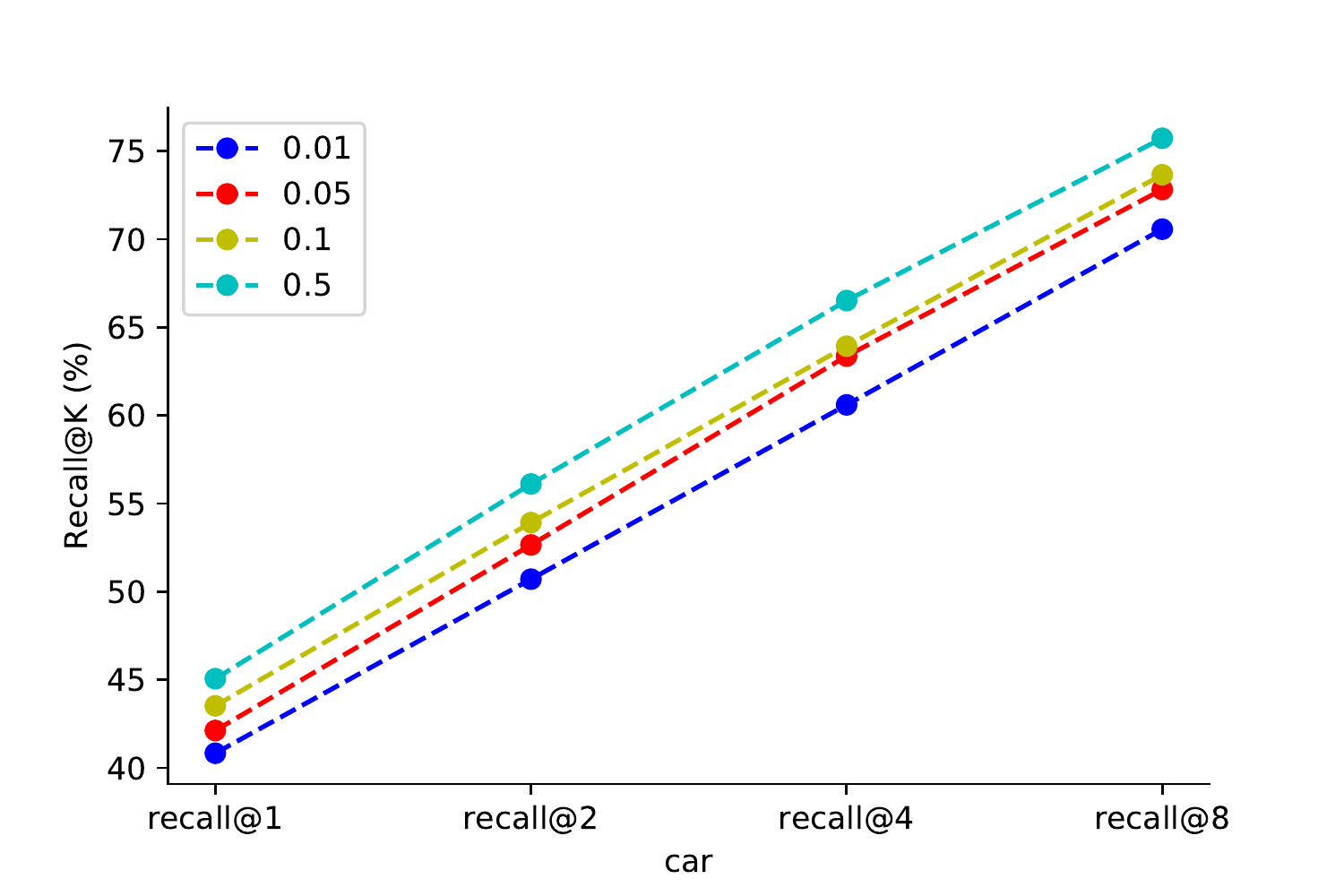}
\end{center}
  \caption{ Performance under different $\eta$ of UDML-SS on Car dataset}
\label{fig:car_etas}
\end{figure}

\begin{figure}[h!]
\begin{center}
\includegraphics[width=0.5\textwidth]{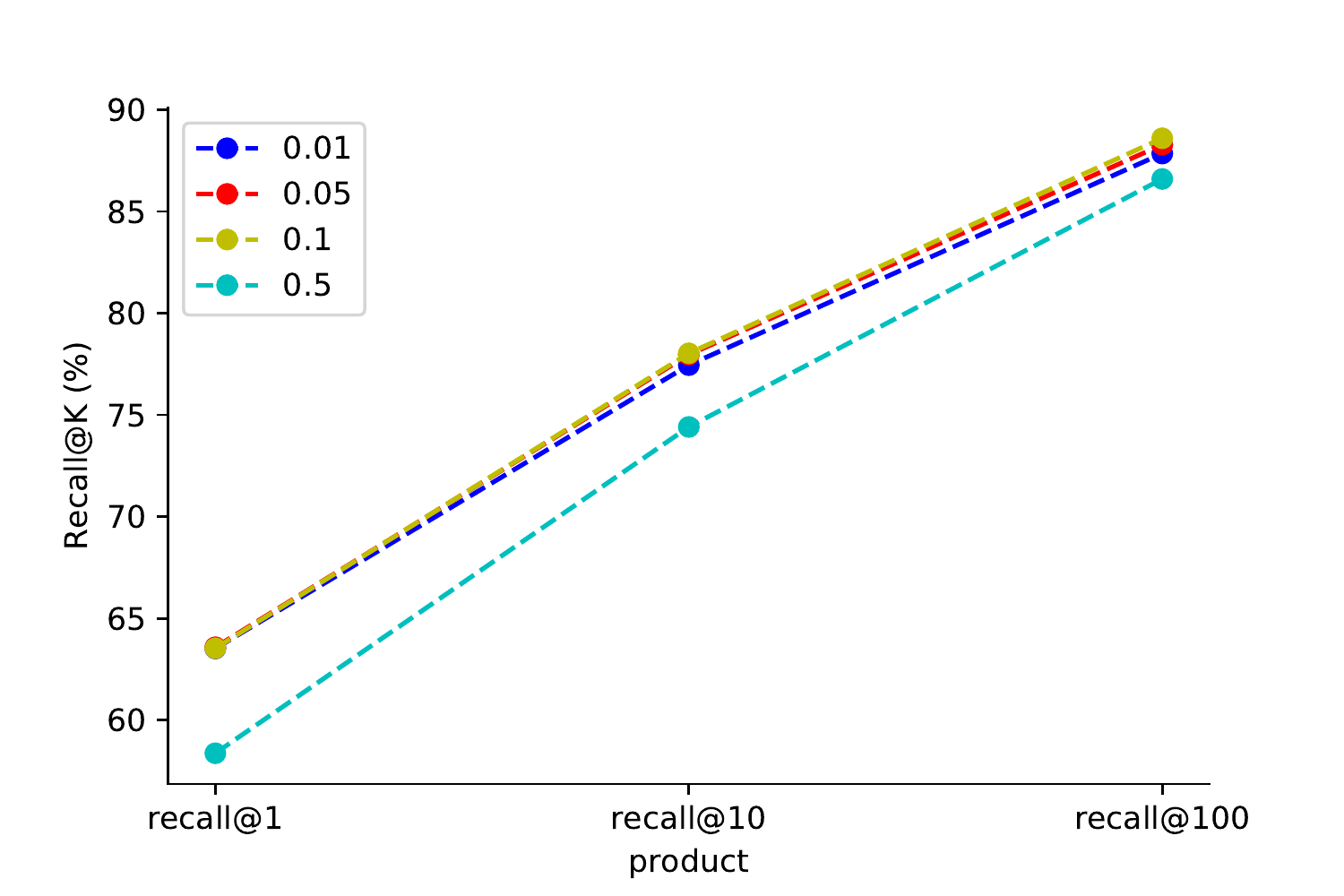}
\end{center}
  \caption{ Performance under different $\eta$ of UDML-SS on Product dataset}
\label{fig:product_etas}
\end{figure}

This section, we show the evaluation results on CUB, Car and Product datasets of UDML-SS using different $\eta$ in Figure \ref{fig:cub_etas}, \ref{fig:car_etas} and \ref{fig:product_etas}. 

\end{document}